\pdfoutput=1

\documentclass[11pt]{article}

\usepackage{acl}

\usepackage{times}
\usepackage{latexsym}
\usepackage{url}
\usepackage{enumitem}
\usepackage{booktabs}
\usepackage{multirow}
\usepackage{array}
\usepackage{graphicx}
\usepackage{subcaption}
\usepackage{amsmath}
\usepackage{amsfonts}
\usepackage{amsthm,amssymb,amsopn,bm}
\usepackage{color,soul}
\usepackage{verbatim}
\usepackage{tabulary}
\usepackage{footnote}
\usepackage{tabularx,booktabs}
\newcolumntype{Y}{>{\centering\arraybackslash}X}
\newcolumntype{P}[1]{>{\centering\arraybackslash}p{#1}}
\usepackage{xspace}
\usepackage{stmaryrd}
\usepackage{makecell}

\usepackage[T1]{fontenc}

\usepackage[utf8]{inputenc}

\usepackage{microtype}
\makesavenoteenv{tabular}
\usepackage{tabularx,booktabs}

\newcommand\bolden[1]{{\boldmath\bfseries#1}}

\newcommand{\dataname}{\textsc{FaVIQ}}
\newcommand{\fever}{\textsc{FEVER}}
\newcommand{\snopes}{\textsc{Snopes}}
\newcommand{\scifact}{\textsc{SciFACT}}
\newcommand{\fm}{\textsc{FM2}}
\newcommand{\boolq}{\textsc{BoolQ-FV}}

\newcommand{\sota}{state-of-the-art}

\newcommand{\N}{R}
\newcommand{\D}{A}
\newcommand{\true}{\texttt{support}}
\newcommand{\false}{\texttt{refute}}
\newcommand{\nei}{{\tt not enough info}}
\newcommand{\neishort}{{\tt NEI}}

\newif\ifmoveanalysis
\moveanalysistrue

\newif\iflongpaper
\longpapertrue

\providecommand{\updated}[1]{
    {\protect\color{red!80!orange}{#1}}
}
\providecommand{\commentout}[1]{}

\newcommand\blfootnote[1]{%
  \begingroup
  \renewcommand\thefootnote{}\footnote{#1}%
  \addtocounter{footnote}{-1}%
  \endgroup
}

\title{\textsc{FaVIQ}: FAct Verification from Information-seeking Questions}

\newcommand{\affilsup}[1]{\rlap{\textsuperscript{\normalfont#1}}}

\author{
    Jungsoo Park\affilsup{1}~\textsuperscript{$*$}
    \qquad Sewon Min\affilsup{2}~\textsuperscript{$*$}
    \qquad Jaewoo Kang\affilsup{1} \\
    \textbf{Luke Zettlemoyer}\affilsup{2}
    \qquad \textbf{Hannaneh Hajishirzi}\affilsup{2,3} \\
    $^1$Korea University \quad $^2$University of Washington \quad $^3$Allen Institute of AI \\
    \texttt{\{jungsoo\_park,kangj\}@korea.ac.kr} \\
    \texttt{\{sewon,lsz,hannaneh\}@cs.washington.edu} \\
}

\begin{document}
\maketitle
\blfootnote{\textsuperscript{$*$}Equal Contribution}
\begin{abstract}
Despite significant interest in developing general purpose fact checking models, it is challenging to construct a large-scale fact verification dataset with realistic real-world claims.
Existing claims are either authored by crowdworkers, thereby introducing subtle biases that are difficult to control for, or manually verified by professional fact checkers, causing them to be expensive and limited in scale.
In this paper, we construct a large-scale challenging fact verification dataset called \dataname, consisting of 188k claims derived from an existing corpus of ambiguous  information-seeking questions. 
The ambiguities in the questions enable automatically constructing true and false claims that 
reflect user confusions (e.g., the year of the movie being filmed vs. being released).
Claims in \dataname\ are verified to be natural, contain little lexical bias, and require a complete understanding of the evidence for verification.
Our experiments show that the state-of-the-art models are far from solving our new task. 
Moreover, training on our data helps in professional fact-checking, outperforming models trained on the widely used dataset FEVER or in-domain data by up to 17\% absolute.
Altogether, our data will serve as a challenging benchmark for natural language understanding and support future progress in professional fact checking.\footnote{Data available at \url{https://faviq.github.io}.}
\end{abstract}

\iflongpaper{
    \section{Introduction}\label{sec:intro}Fact verification, the task of verifying the factuality of the natural language claim, is an important NLP application~\citep{cohen2011computational} and has also been used to evaluate the amount of external knowledge a model has learned~\citep{petroni2020kilt}. However, it is challenging to construct fact verification data with claims that contain realistic and implicit misinformation.
Crowdsourced claims from prior work such as \fever~\citep{thorne2018fever} are written with minimal edits to reference sentences, leading to strong lexical biases such as the overuse of explicit negation 
and unrealistic misinformation that is less likely to occur in real life~\citep{schuster-etal-2019-towards}.
On the other hand, data constructed by professional fact-checkers are expensive and are typically small-scale~\citep{hanselowski2019richly}.

\begin{figure}[t!]
\centering
\includegraphics[width=0.98\linewidth,trim={7cm 6.3cm 7cm 1.5cm},clip]{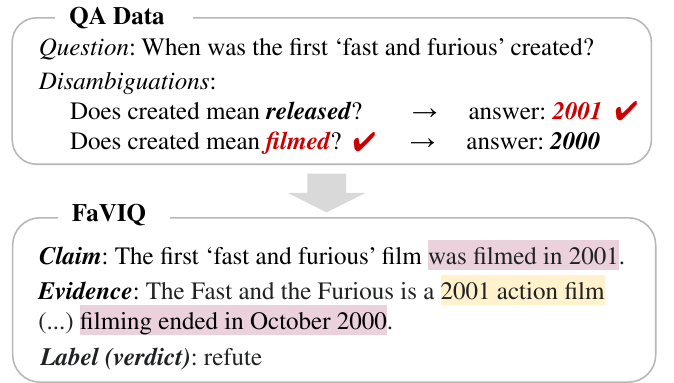}\vspace{-.4em}
\caption{
An example of a \false\ claim on \dataname,
constructed using ambiguity in the information-seeking question, e.g.,
through a crossover of the year of the film {\em being released} and {\em being filmed}.
}
\label{fig:intro}
\end{figure}

In this paper, we show it is possible to use information-seeking questions~\citep{kwiatkowski2019natural} and their ambiguities~\citep{min2020ambigqa} to construct a large-scale, challenging, and realistic fact verification dataset.  
Information-seeking questions are inherently ambiguous because users do not know the answers to the questions they are posing.
For example, in Figure~\ref{fig:intro}, the question is ambiguous because the filming of the movie and the release of the movie can both be seen as the creation time.

We introduce a new dataset
\textbf{\dataname}---\textbf{FA}ct \textbf{V}erification derived from \textbf{I}nformation-seeking \textbf{Q}uestions, which uses such ambiguities to generate challenging fact verification problems. 
For instance, the claim in Figure~\ref{fig:intro} requires the model to identify that the movie released in 2001 is in fact filmed in 2000 and to return \false.
Like this, claims generated through the crossover of the disambiguation of information-seeking questions are likely to contain misinformation that real users are easily confused with.
We automatically generate such claims by composing valid and invalid question-answer pairs and transforming them into textual claims using a neural model.
The data is further augmented by claims from regular question-answer annotations.

\commentout{
We propose to use information-seeking questions~\citep{kwiatkowski2019natural} and their ambiguity~\citep{min2020ambigqa} to construct a large-scale, challenging, and realistic fact verification dataset.  
Information-seeking questions are inherently ambiguous because users do not know the answers to the questions they are posing, which can lead to underspecification or subtle mistakes and misuse of basic facts.

We introduce a new dataset
\textbf{\dataname}---\textbf{FA}ct \textbf{V}erification derived from \textbf{I}nformation-seeking \textbf{Q}uestions, which uses such mistakes to generate challenging fact verification problems. 
Consider the example in Figure~\ref{fig:intro}.
Users ask an ambiguous question because the filming of the movie and the release of the movie are semantically close, and both can be seen as the creation time.
Therefore, claims generated through the crossover of the disambiguation of the information-seeking question are likely to contain misinformation that real users are easily confused with. 
We automatically generate such claims by composing valid and invalid question-answer pairs and transforming them into textual claims using a neural model.
The data is further augmented by claims from regular question-answer annotations.
}

In total, \dataname\ consists of 188k claims.
We manually verified a subset of claims to ensure that they are as natural as human-written claims.
Our analysis shows that the claims have significantly lower lexical bias
than existing crowdsourced claims; claims involve diverse types of distinct entities, events, or properties that are semantically close, being more realistic and harder to verify without a complete understanding of the evidence text.

Our experiments show that a model with no background knowledge performs only slightly better than random guessing, and the state-of-the-art model achieves an accuracy of 65\%, leaving significant room for improvement.
Furthermore, 
training on \dataname\ improves the accuracy of verification of claims written by professional fact checkers,
outperforming models trained on the target data only or pretrained on \fever\ by up to 17\% absolute.
Together, our experiments demonstrate that
\dataname\ is a challenging benchmark as well as a useful resource for professional fact checking.

To summarize, our contributions are three-fold:\vspace{-.1cm}
\begin{enumerate}\setlength\itemsep{-.08cm}
    \item We introduce \dataname, a fact verification dataset consisting of 188k claims.
    By leveraging information-seeking questions and their natural ambiguities, our claims require the identification of entities, events, or properties that are semantically close but distinct, making the fact verification problem very challenging and realistic.
    \item Our experiments show that the state-of-the-art fact verification models are far from solving \dataname, indicating significant room for improvement.
    \item Training on \dataname\ significantly improves the verification of claims written by professional fact checkers, indicating that \dataname\ can support progress in professional fact checking.
\end{enumerate}

    \section{Related Work}\label{sec:related}\begin{figure*}[t]
\resizebox{\textwidth}{!}{\includegraphics{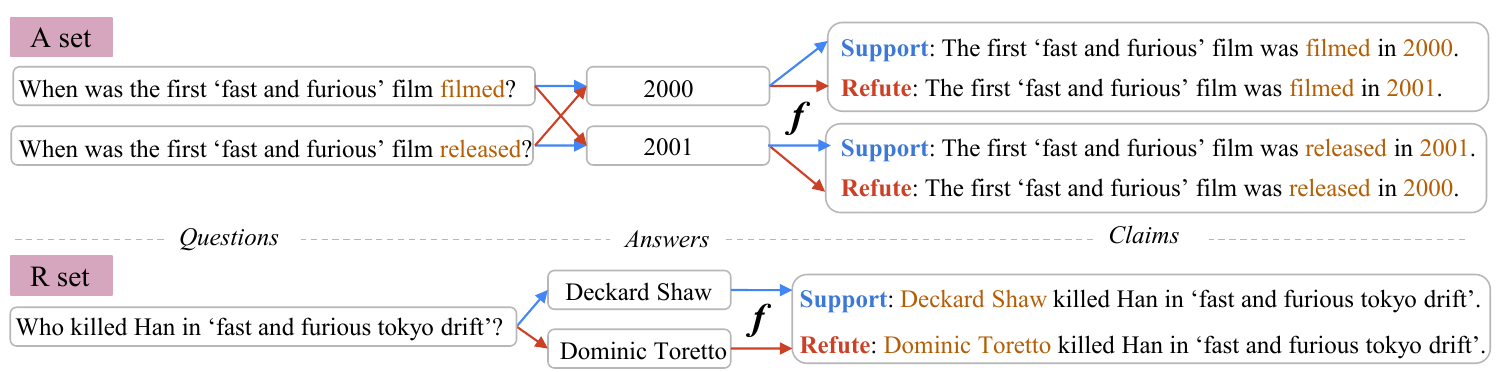}}
\caption{Overview of the data creation process.
The data consits of two sets (\D\ and \N).
For \D, we use the disambiguated question-answer pairs and generate \true\ and \false\ claims from matching pairs ({\em filmed--2000, released--2001}) and crossover pairs ({\em filmed--2001, released--2000}), respectively. For \N, we use the reference answer ({\em Deckard Shaw}) and the incorrect prediction from DPR ({\em Dominic Toretto}) to generate \true\ and \false\ claims, respectively.
$f$ is a T5 model that transforms question-answer pairs to claims (Section~\ref{subsec:qa-to-claim}).
}
\label{fig:process}
\end{figure*}



\paragraph{Fact verification}
Fact verification is crucial for real-world applications~\citep{cohen2011computational} and as a benchmark to evaluate the knowledge in a model~\citep{petroni2020kilt}.

One line of work has studied professional fact checking, dealing with claims collected by professional fact checkers in specific domains
~\citep{vlachos-riedel-2014-fact,ferreira2016emergent,augenstein2019multifc,hanselowski2019richly}.
While such data contains realistic claims that have occurred in the real world,
it is expensive to construct as it requires labor from professional fact checkers.
Moreover,
it is less suitable as a benchmark due to lack of a standard evidence corpus such as Wikipedia\footnote{For this reason, prior work on professional fact checking assumes gold evidence document.} and ambiguities in labels.\footnote{Most claims fall into the \texttt{mixture} label, rather than \true\ or \false.}

Other fact verification datasets are collected through crowdsourcing (e.g., FEVER \citep{thorne2018fever} and its variants \citep{Thorne18Fact, Thorne2019adversarial}) by altering a word or negating the reference text to intentionally make true or false claims. This process leads to large-scale datasets but with strong artifacts and unrealistic claims~\citep{schuster-etal-2019-towards,Thorne2019adversarial,eisenschlos2021fool}.
Consequently, a trivial claim-only baseline with no evidence achieves near 80\% (\citet{petroni2020kilt}, verified in Section~\ref{subsec:baseline-exp}).
While more recent work proposes new crowdsourcing methods that alleviate artifacts~\citep{schuster2021get,eisenschlos2021fool}, their claims are still written given particular evidence text, being vulnerable to subtle lexical biases that can be hard to explicitly measure.

We construct a fact verification dataset from highly ambiguous information-seeking questions. 
Our claims have significantly less lexical bias than other crowdsourced ones (Figure~\ref{fig:lmi_dist}), contain realistic misinformation that people are likely to be confused about (Table~\ref{tab:data-analysis-summary}), and are challenging to current \sota\ models (Section~\ref{subsec:baseline-exp}). 
Moreover, training a model on our data improves professional fact checking (Section~\ref{sec:exp-real}).

\paragraph{QA to Verification Task}
Prior work has also used QA data to create entailment or fact verification benchmarks.
Most make use of synthetic or annotated questions~\citep{demszky2018transforming,jiang2020hover,pan-etal-2021-Zero-shot-FV,chen2021can}\footnote{
    Annotated questions are simulated by crowdworkers given the evidence text and the answer, having largely different distributions from information-seeking questions~\citep{lee2019latent,gardner2019question}.
} 
while we use questions posed by real users to reflect confusions that naturally occur while seeking information.
\citet{thorne2021boolq} use information-seeking questions, by converting yes/no questions to \true/\false\ claims, but at a small scale and with unambiguous questions.
Instead, our work uses large-scale information-seeking questions (with no restriction in answers) to claims.
We are also unique in using highly ambiguous QA pairs to obtain claims that are more challenging to verify and have significantly fewer lexical cues (quantitative comparisons in Section~\ref{subsec:data-analysis}).

    \section{Data}\label{sec:data}\subsection{Data Construction}\label{subsec:data-construction}

We construct \textbf{\dataname}---\textbf{FA}ct \textbf{V}erification derived from \textbf{I}nformation-seeking \textbf{Q}uestions,
where the model is given a natural language claim and predicts \true\ or \false\ with respect to the English Wikipedia.
The key idea to construct the data is to gather a set of valid and invalid question-answer pairs (Section~\ref{subsec:obtain-qa-pairs}) from annotations of information-seeking questions and their ambiguities (Section~\ref{subsec:sources}), and then convert each question-answer pair $(q, a)$ to a claim (Section~\ref{subsec:qa-to-claim}).  
Figure~\ref{fig:process} presents an overview of this process.

\subsubsection{Data Sources}\label{subsec:sources}
We use QA data from Natural Questions (NQ, \citet{kwiatkowski2019natural}) and AmbigQA~\citep{min2020ambigqa}. NQ is a large-scale dataset consisting of the English information-seeking questions mined from Google search queries. AmbigQA provides disambiguated question-answer pairs for NQ questions, thereby highlighting the ambiguity that is inherent in information-seeking questions. 
Given an ambiguous question, it provides a set of multiple distinct answers, each paired with a new disambiguated question that uniquely has that answer. 

\subsubsection{Composing Valid and Invalid QA Pairs}\label{subsec:obtain-qa-pairs}
\dataname\ is constructed from ambiguous questions and their disambiguation ({\em \D} set) and is further augmented by using unambiguous question-answer pairs 
({\em \N} set).

\paragraph{From ambiguous questions ({\em \D} set)}
We use the data consisting of a set of $\left(q, \{q_1, a_1\}, \{q_2, a_2\}\right)$,
where $q$ is an information seeking question that has $a_1, a_2$ as multiple distinct answers.\footnote{
    If $q$ has more than two distinct answers, we sample two. This is to construct a reasonable number of claims per $q$.
}
$q_1$ and $q_2$ are disambiguated questions for the answers $a_1$ and $a_2$, i.e., $q_1$ has $a_1$ as a valid answer and $a_2$ as an invalid answer.
We use $(q_1, a_1)$ and $(q_2, a_2)$ as valid question-answer pairs, and $(q_1, a_2)$ and $(q_2, a_1)$ as invalid question-answer pairs. 

This data is particularly well suited to fact checking because individual examples require identification of entities, events, or properties that are semantically close but distinct: the fact that a user asked an ambiguous question $q$ without realizing the difference between $(q_1, a_1)$ and $(q_2, a_2)$ indicates that the distinction is non-trivial and is hard to notice without sufficient background knowledge about the topic of the question.


\paragraph{From regular questions ({\em \N} set)}
We use the QA data consisting of a set of $\left(q, a \right)$: an information-seeking question $q$ and its answer $a$. We then obtain an invalid answer to $q$, denoted as $a_\mathrm{neg}$, from an off-the-shelf QA model for which we use the model from \citet{karpukhin2020dense}---DPR followed by a span extraction model. We choose $a_\mathrm{neg}$ with heuristics to obtain hard negatives but not the false negative; details provided in Appendix~\ref{app:data-construction}.
We use $(q, a)$ and $(q, a_\mathrm{neg})$ as a valid and an invalid question-answer pair, respectively.

We can think of $(q, a_\mathrm{neg})$ as a {\em hard} negative pair chosen adversarially from the QA model.\footnote{
It is possible that the \N\ set contains bias derived from the use of DPR. We thus consider the \N\ set as a source for data augmentation, while \D\ provides the main data.}
This data can be obtained on a much larger scale than the \D\ set because annotating a single valid answer is easier than annotating disambiguations.

\begin{table}[t]
    \centering \small
    \begin{tabular}{ll rrr}
        \toprule
            && Total & Support & Refute \\
        \midrule
            \multirow{2}{*}{Train}  & \D & 17,008 & 8,504 & 8,504  \\
                                    & \N & 140,977 & 70,131 & 70,846 \\
        \midrule
            \multirow{2}{*}{Dev}    & \D & 4,260  & 2,130 & 2,130  \\
                                    & \N & 15,566  & 7,739 & 7,827  \\
        \midrule
            \multirow{2}{*}{Test}   & \D & 4,688 & 2,344 & 2,344  \\
                                    & \N & 5,877 & 2,922 & 2,955  \\
        \bottomrule
    \end{tabular}
    \caption{\dataname\ statistics.
    {\em \D} includes claims derived from ambiguous questions, while {\em \N} includes claims from regular question-answer pairs.
    }\label{tab:data-statistics}
\end{table}

\subsubsection{Transforming QA pairs to Claims}\label{subsec:qa-to-claim}
We transform question-answer pairs to claims by training a neural model which maps $(q, a)$ to a claim that is \true\ {\em if and only if} $a$ is a valid answer to $q$, otherwise \false.
We first manually convert 250 valid and invalid question-answer pairs obtained through Section~\ref{subsec:obtain-qa-pairs} to claims.
We then train a T5-3B model~\citep{raffel2019exploring}, using 150 claims for training and 100 claims for validation.
The model is additionally pretrained on data from~\citet{demszky2018transforming}, see Appendix~\ref{app:data-construction}. 

\subsubsection{Obtaining silver evidence passages}\label{subsubsec:silver}
We obtain silver evidence passages for \dataname\ by
(1) taking the question that was the source of the claim during the data creation (either a user question from NQ or a disambiguated question from AmbigQA), (2) using it as a query for TF-IDF over the English Wikipedia, and (3) taking the top passage that contains the answer.
Based on our manual verification on 100 random samples, the precision of the silver evidence passages is 70\%.
We provide silver evidence passages primarily for supporting training of the model, and do not explicitly evaluate passage prediction; more discussion in Appendix~\ref{app:data-construction}.
Future work may use human annotations on top of our silver evidence passages in order to further improve the quality, or evaluate passage prediction.

\subsection{Data Validation}\label{subsec:data-validation}
In order to evaluate the quality of claims and labels, three native English speakers were given 300 random samples from \dataname, and were asked to: (1) verify whether the claim is as natural as a human-written claim, with three possible ratings (perfect, minor issues but comprehensible, incomprehensible), and (2) predict the label of the claim (\true\ or \false). Validators were allowed to use search engines, and were encouraged to use the English Wikipedia as a primary source.

Validators found 80.7\% of the \D\ set and 89.3\% of the \N\ set to be natural, and 0\% to be incomprehensible.
The rest have minor grammatical errors or typos, e.g., missing ``the''. In most cases the errors actually come from the original NQ questions which were human-authored, indicating that these grammatical errors and typos occur in real life.
Lastly, validators achieved an accuracy of 95.0\% (92.7\% of \D\ and 97.3\% of \N) when evaluated against gold labels in the data---this indicates high-quality of the data and high human performance.
This accuracy level is slightly higher than that of \fever\ (91.2\%).

\begin{table}[!t]
    \centering \small
    \begin{tabular}{
        l 
        r 
        r 
        r 
        r 
        r
    }
        \toprule
            \multirow{2}{*}{Data} & \multirow{2}{*}{Size} & \multicolumn{4}{c}{Length of the claims} \\
            \cmidrule(lr){3-6}
            & & Avg & Q1 & Q2 & Q3 \\
        \midrule
            \multicolumn{6}{l}{\em Professional claims} \\
            \snopes & 16k & 12.4 & 10 & 11 & 14 \\
            \scifact & 1k & 11.5 & 9 & 11 & 13 \\
        \midrule
            \multicolumn{6}{l}{\em Crowdsourced claims} \\
            \fever & 185k & 9.3 & 7 & 9 & 11 \\
            \fm & 13k & 13.9 & 9 & 13 & 16.3 \\
            \boolq & 10k & 8.7 & 8 & 8 & 9 \\
            \textbf{\dataname} & \textbf{188k} & \textbf{12.0} & \textbf{9} & \textbf{10.5} & \textbf{13.5} \\
        \bottomrule
    \end{tabular}
    \caption{Statistics of a variety of fact verification datasets. 
    {\em Avg} and {\em Q1--3} are the average and quantiles of the length of the claims based on whitespace tokenization on the validation data; for \dataname, we report the macro-average of the \D\ set and the \N\ set.
    dataname\ is as large as \fever\ and has a distribution of claim lengths that is much closer to that of professional fact checking datasets (\snopes\ and \scifact).
    }\label{tab:data-comparison}
\end{table}

\subsection{Data Analysis}\label{subsec:data-analysis}
Data statistics for \dataname\ are listed in Table~\ref{tab:data-statistics}.
It has 188k claims in total, with balanced \true\ and \false\ labels.
We present quantitative and qualitative analyses showing that claims on \dataname\ 
contain much less lexical bias than other crowdsourced datasets and include misinformation that is realistic and harder to identify.

\paragraph{Comparison of size and claim length}
Table~\ref{tab:data-comparison} compares
statistics of 
a variety of fact verification datasets: \snopes~\citep{hanselowski2019richly}, \scifact~\citep{wadden2020fact}, \fever~\citep{thorne2018fever}, \fm~\citep{eisenschlos2021fool}, \boolq~\citep{thorne2021boolq} and \dataname.

\dataname\ is as large-scale as \fever, while its distributions of claim length is much closer to claims authored by professional fact checkers (\snopes\ and \scifact). 
%
%
\fm\ is smaller scale, due to difficulty in scaling multi-player games used for data construction, and has claims that are slightly longer than professional claims, likely because they are intentionally written to be difficult.
\boolq\ is smaller, likely due to relative difficulties in collecting naturally-occurring yes/no questions.

\begin{figure}[t!]
\centering
\includegraphics[width=\linewidth]{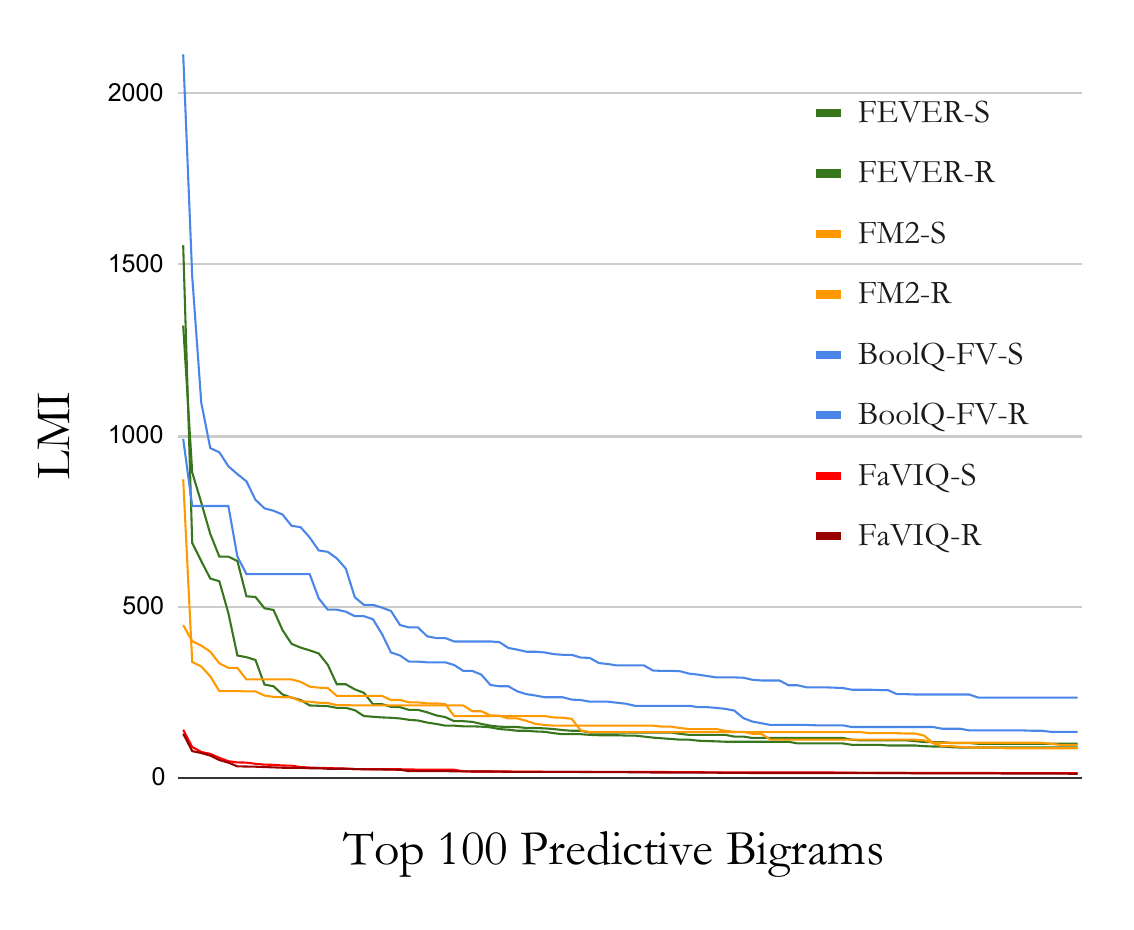}
\caption{
Plot of LMI 
scores of top 100 predictive bigrams for \fever, \fm, \boolq\ and \dataname\ (macro-averaged over the \D\ set and the \N\ set).
\texttt{S} and \texttt{R} denotes \true\ and \false, respectively. \boolq\ indicates data from~\citet{thorne2021boolq} that uses \textsc{BoolQ}.
LMI scores of \dataname\ are significantly lower than those of \fever\ and \fm, indicating significantly less lexical overlap.}
\label{fig:lmi_dist}
\end{figure}

\paragraph{Lexical cues in claims}
We further analyze lexical cues in the claims on \fever, \fm, \boolq\ and \dataname\ by measuring local mutual information (LMI; \citet{schuster-etal-2019-towards, eisenschlos2021fool}).
LMI measures whether the given bigram correlates with a particular label.
More specifically, LMI is defined as:\begin{equation*}
    LMI(w, c) = P(w,c)~log\frac{P(w,c)}{P(w) \cdot P(c)}, \nonumber \label{eq:LMI}
\end{equation*} where $w$ is a bigram, $c$ is a label, and $P(\cdot)$ are estimated by counting~\citep{schuster-etal-2019-towards}.

\begin{table}[t]
    \centering 
    \resizebox{\linewidth}{!}{
    \begin{tabular}{lll}
        \toprule
            Dataset & Top Bigrams by LMI 
            \\
        \midrule
            \fever-\texttt{S} & is a, a film, of the, is an, in the, in a \\
            \fever-\texttt{R} & \hl{is only}, \hl{only a}, \hl{incapable of}, \hl{is not}, \hl{was only}, \hl{is incapable} \\
            \D\ set-\texttt{S} & on the, was the, {the date}, {date of}, {in episode}, is what \\
            \D\ set-\texttt{R} & of the, the country, {at the}, {the episode}, {started in}, placed at \\
            \N\ set-\texttt{S} & {out on}, {on october}, {on june}, {released on}, be 18, {on august} \\
            \N\ set-\texttt{R} & {out in}, {on september}, was 2015, of the, is the, {released in}\\
        \bottomrule
    \end{tabular}}
    \caption{Top bigrams with the highest LMI for \fever\ and \dataname. 
    \texttt{S} and \texttt{R} denotes \true\ and \false\, respectively.
    Highlighted bigrams indicate negative expressions, e.g., ``only'', ``incapable'' or ``not''.
    }\label{tab:data-lmi}
\end{table}

The distributions of the LMI scores for the top-100 bigrams are shown in Figure~\ref{fig:lmi_dist}. The LMI scores of \dataname\ are significantly lower than those of \fever, \fm, and \boolq, indicating that \dataname\ contains significantly less lexical bias.

Tables~\ref{tab:data-lmi} shows the top six bigrams with the highest LMI scores for \fever\ and \dataname.
As highlighted, all of the top bigrams in \false\ claims of \fever\ contain negative expressions, e.g., ``is only'', ``incapable of'', ``did not''.
In contrast, the top bigrams from \dataname\ do not include obvious negations and mostly overlap across different labels, strongly suggesting the task has fewer lexical cues.
Although there are still top bigrams from \dataname\ causing bias (e.g., related to time, such as `on October'), their LMI values are significantly lower compared those from other datasets.
%

\begin{table}[t]
    \centering \footnotesize
    \begin{tabular}{l}
        \toprule
            \textbf{Conjunctions (33.3\%)} \\
            {\em C}: Johannes bell was the foreign minister that signed the \\ treaty of versailles from
            germany. / {\em E}: Johannes bell served \\ as Minister of Colonial Affairs ... He was one of the two \\ German representatives who signed the Treaty of Versailles. \\
            \textbf{Shared attributes (26.7\%)} \\
            {\em C}: Judi bowker played andromeda in the 2012 remake of \\ the 1981 film clash of the titans called wrath of the titans. \\ {\em E}: Judi bowker ... Clash of the Titans (1981). \\
            \textbf{Procedural event (16.7\%)} \\
            {\em C}: Mccrory's originally filed for bankruptcy on february \\ 2002. / {\em E}: McCrory Stores ... by 1992 it filed for bankrupt- \\ cy. ... In February 2002 the company ceased operation. \\
        \midrule
            \textbf{Negation (30.0\%)} \\
            {\em C}: Southpaw hasn't been released yet.
            {\em E}: Southpaw is an \\ American sports drama film released on July 24, 2015. \\
            \textbf{Cannot find potential cause (20.0\%)} \\ 
            {\em C}: Mutiny on the Bounty is Dutch.
            {\em E}: Mutiny on the \\ Bounty is a 1962 American historical drama film. \\
            \textbf{ Antonym (13.3\%)} \\
            {\em C}: Athletics lost the world series in 1989. 
            {\em E}: The 1989 \\ World Series ... with the Athletics sweeping the Giants. \\
        \bottomrule
    \end{tabular}
    \caption{
    Three most common categories based on 30 \false\ claims randomly sampled from the validation set,
    for \dataname\ (top) and \fever\ (bottom) respectively.
    Full statistics and examples in Appendix~\ref{app:data-refute}.
    {\em C} and {\em E} indicate the claim and evidence text, respectively.
    \texttt{Refute} claims in \dataname\ are more challenging, not containing explicit negations or antonyms.
    }\label{tab:data-analysis-summary}
\end{table}

\paragraph{Qualitative analysis of the \false\ claims}

We also analyzed 30 randomly sampled \false\ claims from \dataname\ and \fever\, respectively. We categorized the cause of misinformation as detailed in Appendix~\ref{app:data-refute}, and show three most common categories for each dataset as a summary in Table~\ref{tab:data-analysis-summary}.

On \dataname, 60\% of the claims involve entities, events or properties that are semantically close, but still distinct. For example, they are specified with conjunctions (e.g., ``was foreign minister'' and ``signed the treaty of versailles from germany''), or share key attributes (e.g., films with the same title).
This means that relying on lexical overlap or partially understanding the evidence text would lead to incorrect predictions; one must read the full evidence text to realize that the claim is false.
Furthermore, 16.7\% involve events, e.g., from filing for bankruptcy for the first time to completely ceasing operations (Table~\ref{tab:data-analysis-summary}). This requires full understanding of the underlying event and tracking of state changes~\citep{das2018building,amini2020procedural}.

The same analysis on \fever\ confirms the findings from~\citet{schuster-etal-2019-towards,eisenschlos2021fool}; many of claims contain explicit negations (30\%) and antonyms (13\%), with misinformation that is less likely to occur in the real world (20\%).\footnote{For instance, consider the claim ``Mutiny on the Bounty is Dutch'' in Table~\ref{tab:data-analysis-summary}. There is no Dutch producer, director, writer, actors, or actress in the film---we were not able to find a potential reason that one would believe that the film is Dutch.}


    \section{Experiments}\label{sec:exp-baseline}
We first evaluate \sota\ fact verification models on \dataname\ in order to establish baseline performance levels (Section~\ref{subsec:baseline-exp}).
We then conduct experiments on professional fact-checking datasets to measure the improvements from training on \dataname\ (Section~\ref{sec:exp-real}).

\subsection{
Baseline Experiments
on \dataname
}\label{subsec:baseline-exp}

\subsubsection{Models}\label{subsubsec:models}
We experiment with two settings: a zero-shot setup where models are trained on \fever, and a standard setup where models are trained on \dataname. For \fever, we use the KILT~\citep{petroni2020kilt} version following prior work; we randomly split the official validation set into equally sized validation and test sets, as the official test set is hidden.

All models are based on BART~\citep{lewis2019bart}, a pretrained sequence-to-sequence model which we train to generate either \true\ or \false.
We describe three different variants which differ in their input, along with their accuracy on \fever\ by our own experiments.

\vspace{.5em}
\noindent
\textbf{Claim only BART} takes a claim as the only input. Although this is a trivial baseline, it achieves an accuracy of 79\% on \fever.

\vspace{.5em}
\noindent
\textbf{TF-IDF + BART} takes a concatenation of a claim and $k$ passages retrieved by TF-IDF from \citet{chen2017reading}. It achieves 87\% on \fever. We choose TF-IDF over other sparse retrieval methods like BM25~\citep{robertson2009probabilistic} because \citet{petroni2020kilt} report that TF-IDF outperforms BM25 on \fever.

\vspace{.5em}
\noindent
\textbf{DPR + BART} takes a concatenation of a claim and $k$ passages retrieved by DPR~\citep{karpukhin2020dense}, a dual encoder based model. It is the \sota\ on \fever\ based on \citet{petroni2020kilt} and \citet{maillard2021multi}, achieving an accuracy of 90\%.

\paragraph{Implementation details}
We use the English Wikipedia from 08/01/2019 following KILT~\citep{petroni2020kilt}.
We take the plain text and lists provided by KILT and create a collection of passages where each passage has up to 100 tokens. This results in 26M passages. We set the number of input passages $k$ to 3, following previous work~\citep{petroni2020kilt,maillard2021multi}.
Baselines on \dataname\ are jointly trained on the \D\ set and the \N\ set.

Training DPR requires a positive and a negative passage---a passage that supports and does not support the verdict, respectively. 
We use the silver evidence passage associated with \dataname\ 
as a positive, and the top TF-IDF passage that is not the silver evidence passages as a negative.
More training details are in Appendix~\ref{app:exp-details}.
Experiments are reproducible from \url{https://github.com/faviq/faviq/tree/main/codes}.


\subsubsection{Results}\label{subsubsec:results}


Table~\ref{tab:result} reports results
on \dataname.
%
The overall accuracy of the baselines is low, despite their high performance on \fever.
The zero-shot performance is barely better than random guessing, indicating that the model trained on \fever\ is not able to generalize to our more challenging data.
When the baselines are trained on \dataname, the best model achieves an accuracy of 65\% on the \D\ set, indicating that existing \sota~models do not solve our benchmark.\footnote{
We additionally show and discuss the model trained on \dataname\ and tested on \fever\ in Appendix~\ref{app:additional-experiments}. They achieve non-trivial performance (67\%) although being worse than \fever-trained models that exploit bias in the data.}

\paragraph{Impact of retrieval}
The performance of the claim only baseline that does not use retrieval is almost random on \dataname, while achieving nearly 80\% accuracy on \fever.
This result suggests significantly less bias in the claims, and the relative importance of using background knowledge to solve the task.
When retrieval is used, DPR outperforms TF-IDF, consistent with the finding from~\citet{petroni2020kilt}.

\paragraph{\D\ set~~vs.~~\N\ set}
The performance of the models on the \N\ set is consistently higher than that on the \D\ set by a large margin, implying that claims based on ambiguity arisen from real users are more challenging to verify than claims generated from regular question-answer pairs.
This indicates clearer contrast to prior work that converts regular QA data to declarative sentences~\citep{demszky2018transforming,pan-etal-2021-Zero-shot-FV}.

\begin{table}[t]
    \centering \small
    \setlength{\tabcolsep}{0.4em}
    \begin{tabular}{l @{\hspace{2.4em}} P{0.7cm}P{0.7cm}P{0.7cm}P{0.7cm}}
        \toprule
            \multirow{2}{*}{Model}
            & \multicolumn{2}{c}{Dev} & \multicolumn{2}{c}{Test} \\
            \cmidrule(lr){2-3} \cmidrule(lr){4-5}
            & \D & \N & \D & \N \\
        \midrule
            \multicolumn{5}{l}{\em Training on \fever\ (zero-shot)} \\
            Clain only BART & 51.6 & 51.0 & 51.9 & 51.1 \\
            TF-IDF + BART & 55.8 & 58.5 & 54.4 & 57.2 \\
            DPR + BART & 56.0 & 62.3 & 55.7 & 61.2 \\
        \midrule
            \multicolumn{5}{l}{\em Training on \dataname} \\
            Claim only BART & 51.0 & 59.5 & 51.3 & 59.4 \\
            TF-IDF + BART & 65.1 & 74.2 & 63.0 & 71.2 \\
            DPR + BART & \textbf{66.9} & \textbf{76.8} & \textbf{64.9} & \textbf{74.6} \\
        \bottomrule
    \end{tabular}
    \caption{
    Fact verification accuracy on \dataname.
    DPR + BART achieves the best accuracy; however, there is overall significant room for improvement.
    }\label{tab:result}
\end{table}

\providecommand{\red}[1]{{\protect\color{orange}{#1}}}
\providecommand{\blue}[1]{{\protect\color{blue!80!orange}{#1}}}
\providecommand{\labels}[2]{(\red{#1}; \blue{#2})}
\begin{table*}[t]
    \centering \footnotesize
    \begin{tabular}{
        l r l }
        \toprule
            Category & \% & Example \\
        \midrule
            Retrieval error & 38 &
            \makecell[l]{{\em C}: The american show lie to me ended on january 31, 2011. \labels{Support}{Refute} \\
            {\em E}: Lie to Me ... The second season premiered on September 28, 2009 ... The third season, \\ which had its premiere moved forward to October 4, 2010.} \\
        \midrule
            Events & 28 & 
            \makecell[l]{{\em C}: The bellagio in las vegas opened on may, 1996. \labels{Refute}{Support} \\
            {\em E}: Construction on the Bellagio began in May 1996. ... Bellagio opened on October 15, 1998. 
            } \\
        \midrule
            Evidence not explicit & 18 & \makecell[l]{
            {\em C}: The official order to start building the great wall of china was in 221 bc. 
            (\red{Support}; \\ \blue{Refute})
            {\em E}: The Great Wall of China had been built since the Qin dynasty (221–207 BC).}
            \\
        \midrule
            Multi-hop & 16 &
            \makecell[l]{{\em C}: Seth curry's brother played for davidson in college. \labels{Support}{Refute} \\
            {\em E}: Stephen Curry (...) older brother of current NBA player Seth ... He ultimately chose to \\ attend Davidson College, who had aggressively recruited him from the tenth grade. }
            \\
        \midrule
            Properties & 10 & 
            \makecell[l]{{\em C}: The number of cigarettes in a pack of `export as' brand packs in the usa is 20. (\red{Refute}; \\ \blue{Support})
            {\em E}: In the United States, the quantity of cigarettes in a pack must be at least 20. \\ Certain brands, such as Export As, come in packs of 25.} \\
        \midrule
            Annotation error & 4 & {\em C}: The place winston moved to in still game is finport. \labels{Refute}{Support} \\
        \bottomrule
    \end{tabular}
    \caption{
    Error analysis on 50 samples of the \D\ set of \dataname\ validation data.
    {\em C} and {\em E} indicate the claim and retrieved evidence passages from DPR, respectively.
    \red{Gold} and \blue{blue} indicate gold label and prediction by the model, respectively.
    The total exceeds 100\% as one example may fall into multiple categories.
    }\label{tab:error-analysis}
\end{table*}

\paragraph{Error Analysis}
We randomly sample 50 error cases from DPR + BART on the \D\ set of \dataname\ and categorize them, as shown in Table~\ref{tab:error-analysis}.

\vspace{-.1em}
\begin{itemize}[leftmargin=1.3em]
\setlength\itemsep{-0.1em}
\item {\em Retrieval error} is the most frequent type of errors. DPR typically retrieves a passage with the correct topic (e.g., about ``Lie to Me'') but that is missing more specific information (e.g., the end date).
We think the claim having less lexical overlap with the evidence text leads to low recall@$k$ of the retrieval model ($k=3$).

\item 28\% of error cases involve {\em events}. In particular, 14\% involve procedural events, and 6\% involve distinct events that share similar properties but differ in location or time frame. 

\item In 18\% of error cases, retrieved evidence is {\em valid but not notably explicit}, which is naturally the case for the claims occurring in real life.
\dataname\ has this property likely because it is derived from questions that are gathered independently from the evidence text, unlike prior work~\citep{thorne2018fever,schuster2021get,eisenschlos2021fool} with claims written given the evidence text.

\item 16\% of the failure cases require {\em multi-hop} inference over the evidence.
Claims in this category usually involve procedural events or compositions (e.g. ``is Seth Curry's brother'' and ``played for Davidson in college'').
This indicates that we can construct a substantial portion of claims requiring multi-hop inference without having to make data that artificially encourages such reasoning~\citep{yang2018hotpotqa,jiang2020hover}.

\item Finally, 10\% of the errors were made due to a subtle mismatch in {\em properties}, e.g., in the example in Figure~\ref{tab:error-analysis}, the model makes a decision based on ``required minimum number'' rather than ``exact number'' of a particular brand.
\end{itemize}

    \subsection{Professional Fact Checking Experiments}\label{sec:exp-real}
We use two professional fact-checking datasets.

\vspace{.5em}
\noindent
\textbf{\snopes}~\citep{hanselowski2019richly} consists of 6,422 
claims, 
authored and labeled by professional fact-checkers, gathered from the Snopes website.\footnote{\url{https://www.snopes.com}}
We use the official data split.

\vspace{.5em}
\noindent
\textbf{\scifact}~\citep{wadden2020fact} consists of 1,109 claims based on scientific papers, annotated by domain experts.
As the official test set is hidden, we use the official validation set as the test set, and separate the subset of the training data as the validation set to be an equal size as the test set.

\vspace{.3em}
For both datasets, we merge \nei\ (\neishort) to \false, following prior work that converts the 3-way classification to the 2-way classification~\citep{wang2019superglue, sathe2020automated, petroni2020kilt}.



\subsubsection{Models}
As in Section~\ref{sec:exp-baseline}, all models are based on BART which is given a concatenation of the claim and the evidence text and is trained to generate either \true\ or \false.
For \snopes, the evidence text is given in the original data.
For \scifact, the evidence text is retrieved by TF-IDF over the corpus of abstracts from scientific papers, provided in the original data.
We use TF-IDF over DPR because we found DPR works poorly when the training data is very small.

We consider two settings. In the first setting, we assume the target training data is unavailable and compare the model trained on \fever\ and \dataname\ in a zero-shot setup.
In the second setting, we allow training on the target data and compare the model trained on the target data only and the model with the transfer learning---pretrained on either \fever\ or \dataname\ and finetuned on the target data.

To explore models pretrained on \neishort\ labels, we add a baseline that is trained on a union of the KILT version of \fever\ and \neishort\ data from the original \fever\ from \citet{thorne2018fever}.
For \dataname, we also conduct an ablation that includes the \N\ set only or the \D\ set only.

\paragraph{Implementation details}
When using TF-IDF for \scifact, we use a sentence as a retrieval unit, and retrieve the top 10 sentences, which average length approximates that of 3 passages from Wikipedia.
When using the model trained on either \fever\ or \dataname, we use DPR + BART by default, which gives the best result in Section~\ref{subsec:baseline-exp}.
As an exception, we use TF-IDF + BART on \scifact\  for a more direct comparison with the model trained on the target data only that uses TF-IDF.

When the models trained on \fever\ or \dataname\ are used for professional fact checking,
we find models are poorly calibrated, likely due to a domain shift, as also observed by \citet{kamath2020selective} and \citet{desai2020calibration}.
We therefore use a simplified version of Platt scaling, a post-hoc calibration method~\citep{platt1999probabilistic, guo2017calibration, zhao2021calibrate}. Given normalized probabilities of \true\ and \false, denoted as $p_\texttt{s}$ and $p_\texttt{r}$, modified probabilities $p'_\texttt{s}$ and $p'_\texttt{r}$ are obtained via: \begin{equation*}
    \begin{bmatrix}p'_\texttt{s} \\ p'_\texttt{r}\end{bmatrix}
    = \mathrm{Softmax}\left(
    \begin{bmatrix}p_\texttt{s} + \gamma \\ p_\texttt{r}\end{bmatrix} 
    \right),
\end{equation*} where $-1 < \gamma < 1$ is a hyperparameter tuned on the validation set.


\subsubsection{Results}
Table~\ref{tab:real-fact-checking-result} reports accuracy on professional fact-checking datasets, \snopes\ and \scifact.

\begin{table}[t]
    \centering \small
    \begin{tabular}{l P{1.2cm}P{1.2cm}}
        \toprule
            Training & \snopes & \scifact \\
        \midrule
            Majority only & 60.1 & 58.7 \\
        \midrule
            \multicolumn{3}{l}{\em No target data (zero-shot)} \\
            \fever  & 61.6 & 70.0 \\
            \fever\ w/ \neishort & 63.4 & 73.0  \\
            \dataname  & \textbf{68.2} & \textbf{74.7} \\
            \dataname\ w/o \D\ set & 63.1 & 73.3  \\
            \dataname\ w/o \N\ set & 66.3 & 68.7  \\
        \midrule
            \multicolumn{3}{l}{\em Target data available} \\
            Target only & 80.6 & 62.0 \\
            \fever$\xrightarrow{}$target & 80.6 & 76.7  \\
            \fever\ w/ \neishort $\xrightarrow{}$target & 81.6 & 77.0 \\
            \dataname$\xrightarrow{}$target & \textbf{82.2} & \textbf{79.3} \\
            \dataname\ w/o \D\ set$\xrightarrow{}$target & 81.6 & 78.3 \\
            \dataname\ w/o \N\ set$\xrightarrow{}$target & 81.7 & 76.7 \\
        \bottomrule
    \end{tabular}
    \caption{
    Accuracy on the test set of professional fact-checking datasets.
    Training on \dataname\ significantly improves the accuracy on \snopes\ and \scifact, both in the zero-shot setting and in the transfer learning setting.
    }\label{tab:real-fact-checking-result}
\end{table}

\paragraph{Impact of transfer learning}
We find that transfer learning is effective---pretraining on large, crowdsourced datasets (either \fever\ or \dataname) and finetuning on the target datasets always helps.
Improvements are especially significant on \scifact, likely because its data size is smaller.

Using the target data is still important---models finetuned on the target data outperform zero-shot models by up to 20\%.
This indicates that crowdsourced data cannot completely replace professional fact checking data, but transfer learning from crowdsourced data leads to significantly better professional fact checking performance.

\paragraph{\dataname\ vs. FEVER}
Models that are trained on \dataname\ consistently outperform models trained on \fever, both with and without the target data,
by up to 4.8\% absolute.
This demonstrates that \dataname~is a more effective resource than \fever\ for professional fact-checking.

The model on \fever\ is more competitive when \neishort\ data is included, by up to 3\% absolute.
While the models on \dataname\ outperform models on \fever\ even without \neishort\ data, future work can possibly create \neishort\ data in \dataname\ for further improvement.

\paragraph{Impact of the \D\ set in \dataname}

The performance of the models that use \dataname\ substantially degrades when the \D\ set is excluded.
Moreover, models trained on the \D\ set (without \N\ set) perform moderately well despite its small scale, e.g., on \snopes, achieving the second best performance following the model trained on the full \dataname.
This demonstrates the importance of the \D\ set created based on ambiguity in 
questions.

\snopes\ benefits more from the \D\ set than the \N\ set, while \scifact\ benefits more from the \N\ set than the \D\ set. This is likely because \scifact\ is much smaller-scale (1k claims) and thus benefits more from the larger data like the \N\ set. This suggests that having both the \N\ set and the \D\ set is important for performance.

    \section{Conclusion \& Future Work}\label{sec:concl}We introduced \dataname, a new fact verification dataset derived from ambiguous information-seeking questions. We incorporate facts that real users were unaware of when posing the question, leading to false claims that are more realistic and challenging to identify without fully understanding the context.
Our extensive analysis shows that our data contains significantly less lexical bias than previous fact checking datasets, and include \false\ claims that are challenging and realistic.
Our experiments showed that the \sota\ models are far from solving \dataname, and models trained on \dataname\ lead to improvements in professional fact checking.
Altogether, we believe \dataname\ will serve as a challenging benchmark as well as support future progress in professional fact-checking.

\vspace{.2em}
We suggest future work to improve the \dataname\ model with respect to our analysis of the model prediction in Section~\ref{subsubsec:results},
such as improving retrieval, modeling multi-hop inference, and better distinctions between entities, events and properties.
Moreover, future work may investigate using other aspects of information-seeking questions that reflect facts that users are unaware of or easily confused with.
For example, one can incorporate false presuppositions in questions that arise when users have limited background knowledge~\citep{kim2021linguist}.
As another example, one can explore generating \neishort\ claims by leveraging unanswerable information-seeking questions.
Furthermore, \dataname~can potentially be a challenging benchmark for the claim correction, a task recently studied by~\citet{thorne2021evidence} that requires a model to correct the \false\ claims. 


} \fi

\section*{Acknowledgements}
We thank Dave Wadden, James Thorn and Jinhyuk Lee for discussion and feedback on the paper. We thank James Lee, Skyler Hallinan and Sourojit Ghosh for their help in data validation.
This work was supported by 
NSF IIS-2044660, ONR N00014-18-1-2826, 
an Allen Distinguished Investigator Award, a Sloan Fellowship, National Research Foundation of Korea (NRF-2020R1A2C3010638) and the Ministry of Science and ICT,  Korea, under the ICT Creative Consilience program (IITP-2022-2020-0-01819).

\bibliography{acl}
\bibliographystyle{acl_natbib}

\clearpage
\appendix
\iflongpaper{
    \section{Details in Data Construction}\label{app:data-construction}

\paragraph{Details of obtaining \bolden{$a_\mathrm{neg}$}}
We obtain an invalid answer to the question, denoted as $a_\mathrm{neg}$, using an off-the-shelf QA model, for which we use DPR followed by a span extractor~\citep{karpukhin2020dense}.

The most naive way to obtain $a_\mathrm{neg}$ is to
take the highest scored prediction that is not equal to $a$. We however found such prediction is likely to be a valid answer to $q$, either because it is semantically the same as $a$, or because the ambiguity in the question leads to multiple distinct valid answers.
We therefore use two heuristics that we find greatly reduce such false negatives.
First, instead of taking the top incorrect prediction, we obtain the top $k$ predictions $p_1...p_k$ from the model and randomly sample one from $\{p_1...p_k\} \setminus \{a\}$. We use $k=50$.

Although this is not a fundamental solution to remove false negatives, it significantly alleviates the problem, drastically dropping the portion of false negatives from 14\% to 2\% based on our manual verification on 50 random samples.
Second, we train a neural model that is given a pair of the text and classifies whether they are semantically equivalent or not.
This model is based on T5-large, trained and validated respectively on 150 and 100 pairs of $(a, p_i)$ ($i=1...k$) which we manually label.
We then exclude the predictions in $\{p_1...p_k\}$ which are classified as semantically equivalent to $a$ by the classifier.

\paragraph{QA-to-claim converter}
We use a pretrained sequence-to-sequence model trained on a small number of our own annotations. We first manually write 250 claims given valid or invalid question-answer pairs. We then train a T5-3B model~\citep{raffel2019exploring}, using 150 claims for training and 100 claims for validation. Each question-answer pair is fed into T5 with special tokens {\tt question:} and {\tt answer:}, respectively.

When training, we evaluate on the validation data every epoch and stop training when the validation accuracy does not increase for ten epochs.
The accuracy is measured by the exact match score of the generated and the reference text after normalization, which we found to correlate well with the quality of the generated claims. The final model we train achieves 83\% on the validation data.
At inference time, we filter claims that do not contain the answer string, which may happen when the question is overly specific.

\paragraph{Why don't we evaluate evidence prediction?}
Unlike FEVER~\citep{thorne2018fever}, which includes  evidence prediction as part of the task, our paper does not report the evidence prediction performance and mainly reports the classification accuracy. There are three reasons for this change:

\vspace{-.1em}

\begin{enumerate}[leftmargin=1em]
\setlength\itemsep{-0.1em}
\item As claims on \dataname\ were written independent from any reference text, gold evidence text must be gathered through a separate process, which greatly increases the cost. This is different from other annotated fact checking datasets where a crowdworker wrote a claim based on the reference text and therefore the same reference text can be considered as gold evidence.
\item Finding gold evidence text is an inherently incomplete process; no human can get close to, or even measure the upperbound.
Therefore, even after exhaustive human annotation, evaluation against annotated evidence leads to significant amount of false negatives.
For example, when manually evaluating the top negatives of TF-IDF on 50 random samples from \fever, 42\% are false negatives.
\item Including evidence prediction as part of evaluation significantly restricts the approach models can take. For instance, one may choose not to use the text corpus provided in the dataset (e.g., Wikipedia), and decide to use other sources such as structured data (e.g. knowledge bases) or implicit knowledge stored in large neural models.
\end{enumerate}
Nonetheless, as described in Section~\ref{subsubsec:silver}, we still provide the silver evidence passages which is useful to train a model, e.g., DPR, and supports future work to evaluate the evidence prediction accuracy.


\section{Analysis of \false\ Claims}\label{app:data-refute}





\begin{table*}[t]
    \centering \footnotesize
    \begin{tabular}{
        l
        @{\hspace{-0.1\tabcolsep}}
        r
        @{\hspace{0.2\tabcolsep}}
        r
        @{\hspace{1.5\tabcolsep}}
        l }
        \toprule
            Category & \makecell[r]{\% \\ {\scriptsize \dataname}} & \makecell[r]{\% \\ {\scriptsize \fever}} & Example \\
        \midrule
            Negation & 0 & 30.0 & \makecell[l]{
            {\em C}: Southpaw hasn't been released yet. (from \fever) \\
            {\em E}: Southpaw is a 2015 American sports drama film ... released on July 24, 2015.}\\
        \midrule
            Antonym & 3.3 & 13.3 & \makecell[l]{
            {\em C}: Athletics lost the world series in 1989. \\
            {\em E}: The 1989 World Series was ... with the Athletics sweeping the Giants in four games.} \\
        \midrule
            \makecell[l]{Requires reading \\ across conjunctions} & 33.3 & 6.6 & \makecell[l]{
            {\em C}: Johannes bell was the foreign minister that signed the treaty of versailles from \\
            germany. / {\em E}: Johannes bell served as Minister of Colonial Affairs ... He was one of the \\ two German representatives who signed the Treaty of Versailles.
            } \\
        \midrule
            \makecell[l]{Shared attributes} & 26.7 & 6.6 & \makecell[l]{{\em C}: Judi bowker played andromeda in the 2012 remake of the 1981 film clash of the \\ titans called wrath of the titans. / {\em E}: Judi bowker ... Clash of the Titans (1981). 
            }\\
        \midrule
            Procedural event & 16.7 & 0 & \makecell[l]{{\em C}: Mccrory's originally filed for bankruptcy on february 2002. /
            {\em E}: McCrory Stores ... \\ by 1992 it filed for bankruptcy. ... In February 2002 the company ceased operation.
            } \\
        \midrule
            \makecell[l]{Incorrect type of \\ properties} & 10.0 & 3.3 &  \makecell[l]{
            {\em C}: Tyler, the Creator is the name of the song at the end of who dat boy. \\
            {\em E}: "Who Dat Boy" is a song by American rapper Tyler, the Creator. }
            \\
        \midrule
            \makecell[l]{Cannot find potential \\ cause} & 0 & 20.0 & \makecell[l]{{\em C}: Mutiny on the Bounty is Dutch. (from \fever) \\
            {\em E}: Mutiny on the Bounty is a 1962 American Technicolor epic historical drama film.  
            }\\
        \midrule
            Annotation error & 10.0 & 20.0 & \makecell[l]{{\em C}: Pasek and paul were the individuals that wrote the lyrics to the greatest showman.
            }\\
        \bottomrule
    \end{tabular}
    \caption{Categorization of 30 \false\ claims on \dataname\ and \fever, randomly
    sampled from the validation set.
    {\em C} and {\em E} indicate the claim and evidence text, respectively.
    Examples are from \dataname\ unless otherwise specified.
    }\label{tab:data-analysis}
\end{table*}

We randomly sample 30 \false\ claims from \dataname\ and \fever, respectively, and categorize the cause of the misinformation, as shown in Table~\ref{tab:data-analysis}.
See Section~\ref{subsec:data-analysis} for discussion.

\section{Details of Experiments}\label{app:exp-details}

\begin{table}[t]
    \centering \small
    \setlength{\tabcolsep}{0.4em}
    \begin{tabular}{l @{\hspace{2.4em}} P{1.4cm}P{1.7cm}}
        \toprule
            Model & Dev & Test \\
        \midrule
            Clain only BART & 47.2 & 48.3  \\
            TF-IDF + BART & 67.8 & 66.6  \\
            DPR + BART & 67.2 & 66.5 \\
        \bottomrule
    \end{tabular}
    \caption{
    Fact verification accuracy on \fever\ of different models when trained on \dataname.
    }\label{tab:result_fever}
\end{table}

\paragraph{DPR training for \fever}
As \fever\ provides the annotated evidence passage, we use it as a positive training example.
We obtain a negative by querying the claim to TF-IDF and taking the passage that is not the positive passage and has the second highest score.
We initially considered using the negative with the highest score, but found that many of them (37\%) are false negatives based on our manual evaluation of 30 random samples. This is likely due to incomprehensive evidence annotation as discussed in Appendix~\ref{app:data-construction}.
We find using the negative with the second highest instead decreases the portion of false negatives from 37\% to 13\%.

\commentout{
    \paragraph{Calibration}
    When the models trained on \fever\ or \dataname\ are used for professional fact checking,
    we find models are poorly calibrated, likely due to a domain shift, as also observed in \citet{kamath2020selective,desai2020calibration}.
    We therefore use a simplified version of Platt scaling, a post-hoc calibration method~\citep{platt1999probabilistic, guo2017calibration, zhao2021calibrate}. Given normalized probabilities of \true\ and \false, denoted as $p_\texttt{s}$ and $p_\texttt{r}$, modified probabilities $p'_\texttt{s}$ and $p'_\texttt{r}$ are obtained via: \begin{equation*}
        \begin{bmatrix}p'_\texttt{s} \\ p'_\texttt{r}\end{bmatrix}
        = \mathrm{Softmax}\left(
        \begin{bmatrix}p_\texttt{s} \\ p_\texttt{r}\end{bmatrix} + 
        \begin{bmatrix}\gamma \\ 0 \end{bmatrix}
        \right),
    \end{equation*} where $-1 < \gamma < 1$ is a hyperparameter tuned on the validation set.
}
\paragraph{Other details}
Our implementations are based on PyTorch\footnote{\url{https://pytorch.org/}}~\citep{paszke2019pytorch} and Huggingface Transformers\footnote{\url{https://github.com/huggingface/transformers}}~\citep{wolf-etal-2020-transformers}.

When training a BART-based model, we map \true\ and \false\ labels to the words \lq true\rq\ and \lq false\rq\ respectively so that each label is mapped to a single token. This choice was made against mapping to \lq support\rq\ and \lq refute\rq
because the BART tokenizer maps \lq refute\rq\ into two tokens, making it difficult to compare probabilities of \true\ and \false.

By default, we use a batch size of 32, a maximum sequence length of 1024, and 500 warmup steps using eight 32GB GPUs.
For \scifact, we use a batch size of 8 and no warmup steps using four 32G GPUs.
We tune the learning rate in between \{7e-6, 8e-6, 9e-6, 1e-5\} on the validation data.

\section{Additional Experiments}\label{app:additional-experiments}

Table~\ref{tab:result_fever} reports the model performance when trained on \dataname\ and tested on \fever.
The best-performing model achieves non-trivial performance (67\%).
However, their overall performance is not as good as model performance when trained on \fever, likely because the models do not exploit the bias in the \fever\ dataset. Nonetheless, we underweight the test performance on \fever\ due to known bias in the data.

\commentout{

\section{Discussions \& Future Work}\label{app:future-work}


Future work can explore developing retrieval models which are more robust to lexical distractions.
One potential approach is to decompose the claim and verify each segment, as explored by prior work in the context of QA~\citep{lewis2018generative,min2019multi}.
This approach is expected to be particularly effective for claims involving multi-hop inference, conjunctions and procedural events, which are the main sources of the challenging claims (Section~\ref{subsec:data-analysis}; Table~\ref{tab:data-analysis}) and errors made by the model (Table~\ref{tab:error-analysis}).
Such an approach in fact mimics the process we did for obtaining a positive passage to train the DPR model, where we decompose the claim into the question and the answer, use one for querying the TF-IDF index and the other for reranking the candidate positive passages---an approach that is only possible at training time rather than test time.
Moreover, future work can make a connection between the open-domain QA and fact verification models to jointly improve the performance in two tasks, as explored by \citet{chen2021can} in the context of reading comprehension and the entailment task.


}
}\fi

\end{document}